%% file: main.tex
\documentclass[runningheads]{llncs}

\usepackage{eccv}

\usepackage{eccvabbrv}

\usepackage{graphicx}
\usepackage{booktabs}
\usepackage{colortbl}
\usepackage{multirow}
\usepackage{threeparttable}
\usepackage{pifont} 
\usepackage{wrapfig}
\usepackage[accsupp]{axessibility}  % Improves PDF readability for those with disabilities.

\newcommand{\ModelName}{{X-Pose}\xspace}
\newcommand{\DataName}{{UniKPT}\xspace}

\makeatletter
\apptocmd\@maketitle{{\myfigure{}\par}}{}{}
\makeatother
\usepackage[pagebackref,breaklinks,colorlinks,citecolor=eccvblue]{hyperref}
\usepackage{orcidlink}

\begin{document}

% ---------------------------------------------------------------
% TODO REVIEW: Replace with your title
\title{X-Pose: Detecting Any Keypoints} 

% TODO REVIEW: If the paper title is too long for the running head, you can set
% an abbreviated paper title here. If not, comment out.
\titlerunning{X-Pose}

% TODO FINAL: Replace with your author list. 
% Include the authors' OCRID for the camera-ready version, if at all possible.
{\author{Jie Yang\textsuperscript{1,2} \and%
Ailing Zeng\textsuperscript{1,$\star$} \and %
Ruimao Zhang\textsuperscript{2,$\star$} % 
\and
Lei Zhang\textsuperscript{1}}}

\authorrunning{J.~Yang, et al.}

\institute{\textsuperscript{1}IDEA~~~\textsuperscript{2}The Chinese University of Hong Kong, Shenzhen
\\
\url{https://github.com/IDEA-Research/X-Pose}}

\def\thefootnote{}\footnotetext{\textsuperscript{$\star$}Corresponding author.}

\newcommand{\myPara}[1]{\vspace{.05in}\noindent\textbf{#1}}
\newcommand\myfigure{%
\centering
\includegraphics[width=1\textwidth]{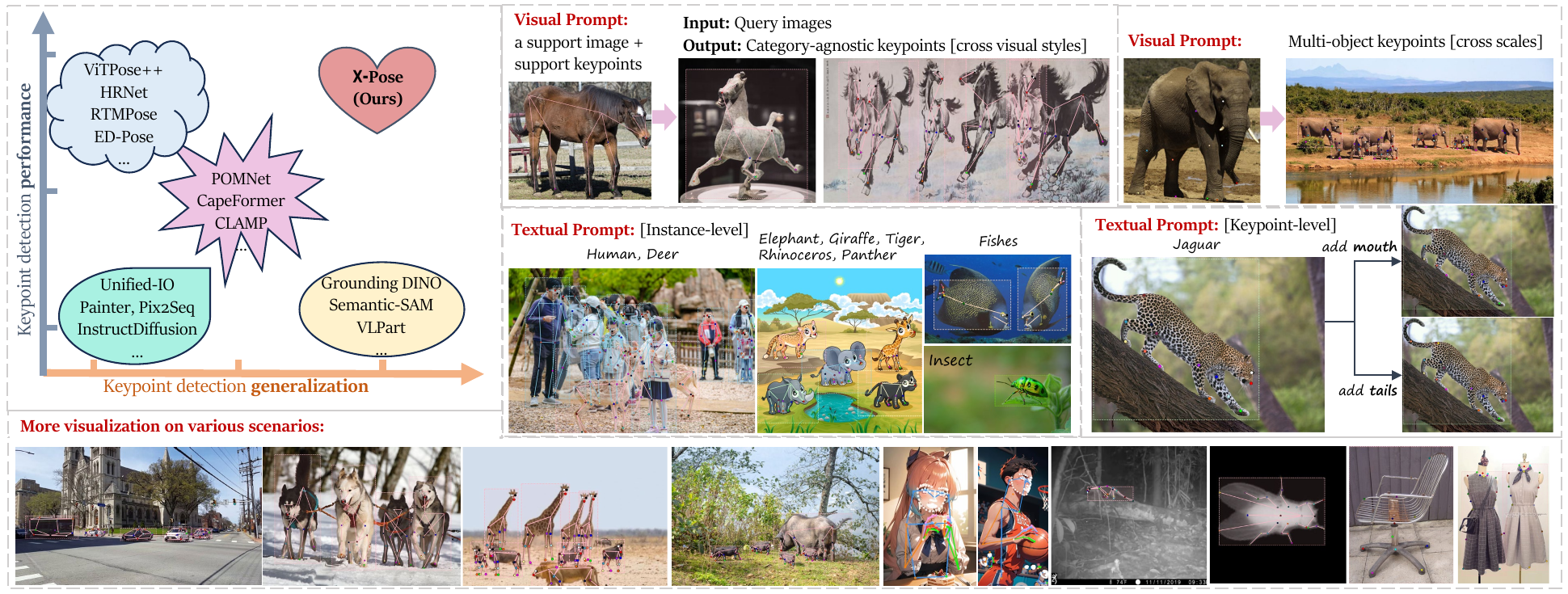}
\vspace{-0.7cm}
\captionof{figure}
{
The proposed \ModelName achieves both strong keypoint detection generalization ability and high performance. 
\ModelName utilizes visual and textual prompts for training to learn fine-grained local-region visual representation via keypoint-text and keypoint-image alignments. Once trained, it can generalize cross-object and cross-keypoint categories, where it can detect multi-object keypoints on various challenging scenarios with diverse visual styles, scales, and poses.}
\label{fig:teaser}
% \vspace{-0.4cm}
}
\maketitle

\begin{abstract}
 This work aims to address an advanced keypoint detection problem: how to accurately detect \textbf{any} keypoints in complex real-world scenarios, which involves massive, messy, and open-ended objects as well as their associated keypoints definitions. Current high-performance keypoint detectors often fail to tackle this problem due to their two-stage schemes, under-explored prompt designs, and limited training data. To bridge the gap, we propose \ModelName, a novel end-to-end framework with multi-modal (i.e., visual, textual, or their combinations) prompts to detect multi-object keypoints for any articulated (e.g., human and animal), rigid, and soft objects within a given image. Moreover, we introduce a large-scale dataset called \DataName, which unifies $13$ keypoint detection datasets with $338$ keypoints across $1,237$ categories over $400$K instances. 
 Training with \DataName, \ModelName effectively aligns text-to-keypoint and image-to-keypoint due to the mutual enhancement of multi-modal prompts based on cross-modality contrastive learning. Our experimental results demonstrate that \ModelName achieves notable improvements of $27.7$ AP, $6.44$ PCK, and $7.0$ AP compared to state-of-the-art non-promptable, visual prompt-based, and textual prompt-based methods in each respective fair setting.
 More importantly, the in-the-wild test demonstrates \ModelName's strong fine-grained keypoint localization and generalization abilities across image styles, object categories, and poses, paving a new path to multi-object keypoint detection in real applications.
 
\end{abstract}
\input{Sec/intro}

\input{Sec/related_work}

\input{Sec/Method}

\input{Sec/data}

\input{Sec/experiment}

\input{Sec/conclusion}
\bibliographystyle{splncs04}
\bibliography{main}

\end{document}

%% file: Sec/intro.tex
\section{Introduction}

\begin{figure*}[h]	
\vspace{-0.5cm}
\centering
 	{
        \includegraphics[width=1.0\linewidth]{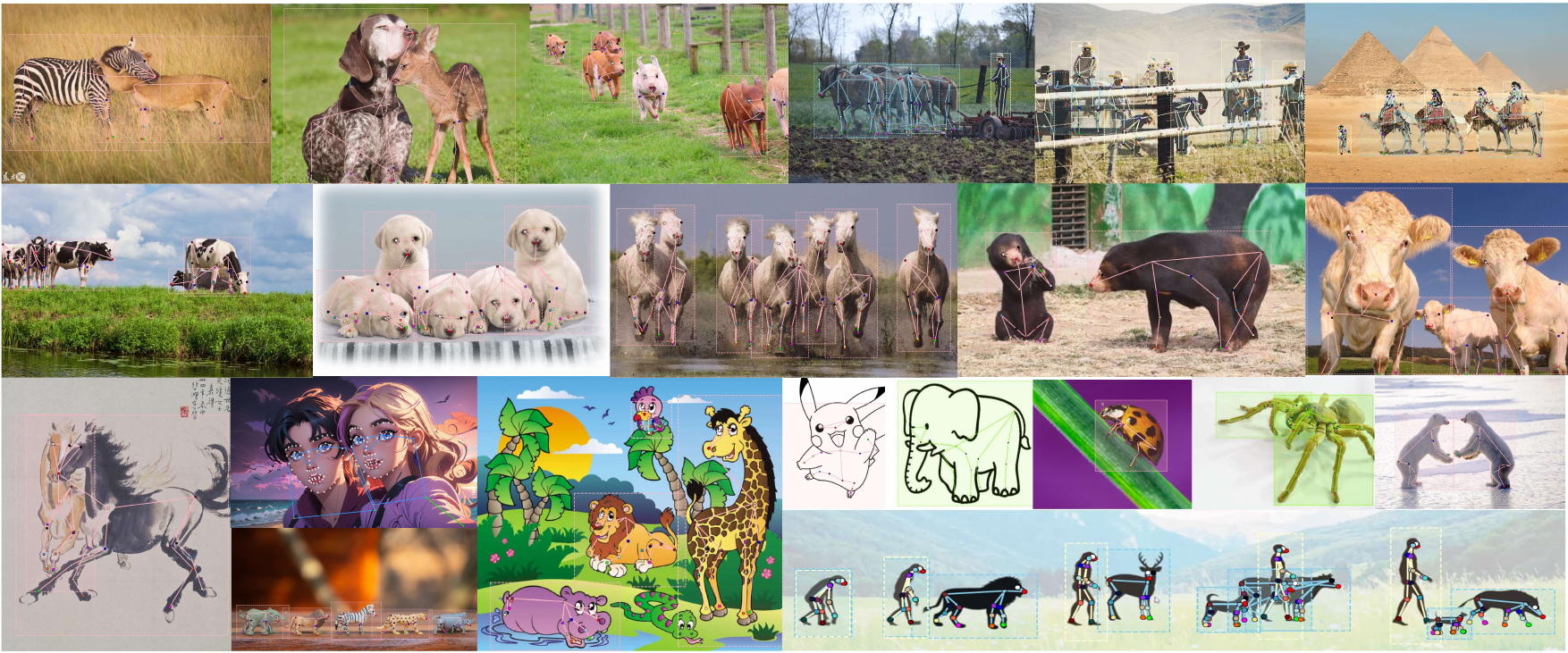}   
 	} 
\vspace{-0.5cm}
\caption{In-the-wild test of \ModelName for any keypoint detection. We highlight the powerful detection performance from cross-category (the first row), multi-object (the second row), and cross-image-style (the third row) with various pose scenarios.}
\label{fig:vis_ours} 
\vspace{-0.5cm}
\end{figure*}

Multi-object keypoint detection, also known as multi-object pose estimation, stands as a fundamental computer vision task with board applications in VR/AR, biomedicine, and robots. It aims to
estimate the 2D keypoint positions of various objects within an image, as shown in Fig.~\ref{fig:teaser} and Fig.~\ref{fig:vis_ours}. With the rapid development of research areas, technologies are gradually transitioning from closed-set to open-world scenarios, leading to the formulation of three model paradigms:
non-promptable model~\cite{xu2022vitpose,jiang2023rtmpose,yang2022explicit,cheng2020higherhrnet}, visual prompt-based model~\cite{xu2022pose,shi2023matching,ge2021metacloth,lu2022few,he2023few} and textual prompt-based model~\cite{zhang2023clamp}.
Although numerous models have been proposed, detecting any keypoints in complex real-world scenes remains a significant challenge due to the following difficulties:

\noindent \textbf{The Bottleneck of Two-stage Schemes.} Current high-performance non-promptable and promptable keypoint detectors all adopt two-stage strategies (e.g., top-down). Given an image containing multiple objects, they require an additional object detector to obtain object bounding boxes. Following this, they can crop the original image into single-object images and proceed with single-object keypoint detection. Consequently, the effectiveness of these models heavily relies on the performance of the object detector.
Despite utilizing state-of-the-art object detectors~\cite{liu2023grounding,zhang2022dino}, challenges persist, 
including inaccurate bounding boxes, missed detections, and excessive redundant bounding boxes due to inaccurate confidence scores, making them less reliable for real-world applications.

\noindent \textbf{Under-explored Prompt Designs.} 
Existing promptable keypoint detectors focus on single-object scenes with only keypoint-level prompts, overlooking object-level prompts. As a result, when faced with unseen objects with significant variations in appearance and different keypoint definitions, the model's ability to generalize at the keypoint level will be severely limited. Furthermore, these methods only accommodate a single type of prompt, either textual or visual, making user interactions unfriendly and inefficient.

\begin{table}[t]
   	\caption{Representative Model Comparisons of different paradigms. TD and E2E are the top-down and end-to-end methods. $N$ is the object number in the image. \textbf{T.} and \textbf{V.} in the first row mean Textual Prompts and Visual Prompts, respectively. 
}
\vspace{-1cm}
	\begin{center}
\resizebox{\linewidth}{!}{
  \begin{threeparttable}
		\begin{tabular}{l|cccc|ccc|c}
			\hline
			Methods & \textbf{T.} & \textbf{V.} & Multi-object &  Multi-class & {Keypoints} & {Object}  & Training Images & Time [ms] \\ \hline
                   \multicolumn{9}{l}{\cellcolor{Gray!25} \textit{ Non-promptable Methods}} \\
            ViTPose~\cite{xu2022vitpose} (TD) & \ding{55} & \ding{55} & \ding{55} & \ding{55} & 17 & 1 & 58K & 60 $\times$ N\\
                        ED-Pose~\cite{yang2022explicit} (E2E) & \ding{55} & \ding{55} & $\checkmark$ & \ding{55}& 17 & 1 & 58K & 55 \\ \hline
                                           \multicolumn{9}{l}{\cellcolor{Gray!25} \textit{ Promptable Methods}} \\
            Capeformer~\cite{shi2023matching} (TD) & \ding{55} & $\checkmark$  & \ding{55} & \ding{55} & 293 & 100 & 17K & 57 $\times$ N \\
            CLAMP~\cite{zhang2023clamp} (TD) &  $\checkmark$ & \ding{55} & \ding{55} & \ding{55}& 14 & 54  & 10K & 63 $\times$ N\\ 
            \ModelName (E2E) & $\checkmark$ & $\checkmark$  & $\checkmark$& $\checkmark$ & \textbf{338} & \textbf{1237} & \textbf{226K} &  59 \\ \hline
		\end{tabular}
      \end{threeparttable}
		}
   \label{tab:comp}
	\end{center}
  \vspace{-0.7cm}
\end{table}

\noindent \textbf{Limited Training Data.} Existing textual-prompt keypoint detectors~\cite{zhang2023clamp} are trained using only $17$ keypoint descriptions for animal pose estimation~\cite{yu2021ap}. Similarly, present visual-prompt keypoint detectors~\cite{xu2022pose,shi2023matching} rely on a small-scale dataset~\cite{xu2022pose} ({e.g.}, only 20K images with 100 instance classes). The constrained nature of their training data significantly hampers generalizability and effectiveness.
Currently, there is no dataset encompassing a broader range of object categories, a more extensive set of keypoint categories, and a larger quantity of data. Organizing such a dataset would facilitate the learning of structured keypoint representations across various categories, thereby bolstering the models' generalization in real-world applications.

Considering the above challenges and motivations, we present \ModelName, a fully end-to-end framework that leverages multi-modal prompts (\textit{i.e.,} textual, visual, or their combinations) to detect multi-object keypoints in complex real-world scenarios.
\noindent \textbf{Firstly}, inspired by the DETR-like non-promptable human pose estimator~\cite{yang2022explicit}, which integrates human-level and keypoint-level detection into an end-to-end framework,
we extend this framework by incorporating the prompt mechanism to support any objects and keypoints.
Such a scheme could dynamically adjust the object categories to be detected and accommodate diverse keypoint definitions, making the model effectively generalize across multi-class and multi-object scenarios in an open-world setting.
\textbf{Secondly}, \ModelName is the first model to leverage multi-modal prompts for multi-object keypoint detection. 
 During training, incorporating textual prompts into visual prompt-based keypoint detection tasks provides valuable high-level semantic guidance, enhancing the learning process. Similarly, visual prompts can complement textual prompts by providing detailed image-level information, thereby improving the effectiveness of both tasks. Moreover, during inference, a prompt design that supports textual, visual, and combination prompts can significantly enhance user experience and efficiency.
\textbf{Lastly}, for effectively training \ModelName, we present {UniKPT}, a large-scale dataset that unifies 13 keypoint detection datasets with 338 keypoints across 1,237 categories over 400K instances. 
In {UniKPT}, we balance these datasets by considering image appearance and style diversity, instances with varying poses, viewpoints, visibilities, and scales. 
Additionally, we reconstruct the semantic relationships between all keypoints and categories, while also standardizing the definitions of the same keypoints across categories (e.g., the left eye of different animals). Most importantly, each keypoint is meticulously annotated with a unique name to enhance keypoint-level semantic understanding.

Based on the above technical and data contributions, we show the functionalities supported and efficiency comparison with the previous representative models in Tab.~\ref{tab:comp}. Furthermore, 
through comprehensive experiments, we demonstrate the remarkable generalization capabilities of \ModelName for both visual prompt-based and textual prompt-based keypoint detection. Compared to state-of-the-art methods, it achieves notable improvements of \textbf{6.44} PCK and \textbf{7.0} AP. Moreover, \ModelName significantly outperforms the state-of-the-art end-to-end model, particularly achieving a \textbf{27.7} AP improvement in AP-10K~\cite{yu2021ap}. Its performance is also comparable to state-of-the-art results on all existing keypoint datasets.
Additionally, \ModelName exhibits impressive text-to-image alignment at both object and keypoint levels, surpassing CLIP by \textbf{204}\% when distinguishing between different animal categories and by \textbf{166}\% across various image styles. %From Fig.~\ref{fig:fig2}-(d), 
From extensive qualitative results on in-the-wild images, we showcase the open-world keypoint detection performance and generalization ability of \ModelName, hoping it could benefit fine-grained visual perception and understanding.

%% file: Sec/related_work.tex
\section{Related Work}
This section introduces three related areas, including non-promptable, visual prompt-based, and textual prompt-based keypoint detection.
As summarized in Tab.~\ref{tab:comp}, our \ModelName is the \textbf{first} end-to-end multi-modal prompt-based model, which could effectively and efficiently detect any keypoints in complex real-world scenarios involving multi-classes and multi-objects as well as their associated keypoints definitions.

\subsection{Non-Promptable Keypoint Detection}
Existing non-promptable methods have mainly focused on human or animal pose estimation~\cite{sun2023uniap,ye2022superanimal,Mathisetal2018,ng2022animal,xu2022vitpose,xu2022vitpose+,jiang2023rtmpose,yang2022explicit,Zhou_2023_ICCV}, categorized into two-stage and one-stage paradigms.
Among the two-stage methods, the top-down strategies dominate and demonstrate high performance~\cite{xiao2018simple,sun2019deep,li2021pose,mao2022poseur,xu2022vitpose,geng2023human} by first detecting each object in an image using an independent object detector and then solely focusing on single-object keypoint detection with the proposed model. However, the use of separate object detectors and multiple inferences for each object incurs high computational costs. Moreover, any missed object detections directly lead to failures in keypoint detection.
Recently, one-stage methods~\cite{shi2022end,yang2022explicit,liu2023group,yang2023neural} have proposed to detect multi-person keypoints in an end-to-end manner, which have shown superior performance and efficiency trade-offs.
However, these methods only focus on single-class objects with specific pre-defined keypoints, limiting their applicability and generality in real-world scenes.
In light of these, our work aims to provide an end-to-end keypoint detector with strong keypoint generalization to detect any keypoints in complex real-world scenes.

\subsection{Visual Prompt-based Keypoint Detection}
Given a prompt image of a novel object and its corresponding keypoint definitions, visual prompt-based keypoint detection aims to detect the keypoints of the same object within an image.
Existing methods~\cite{xu2022pose,shi2023matching,ge2021metacloth,he2023few,lu2022few,sun2023uniap,ye2022superanimal,lauer2022multi} typically focus on single-object scenes, simplifying this problem. Therefore, these methods are unable to address multi-class multi-object scenarios without known object detection, particularly situations where an image contains numerous objects of different categories with varying keypoint definitions. Moreover, most of them only consider one super-category, such as clothing or animal. To handle more objects, recent works~\cite{xu2022pose,shi2023matching} train their models on the MP-100 dataset with $17$K images spanning $100$ categories. However, such a small-scale dataset makes them suffer from under-fitting and hard to learn the local keypoint representation effectively.
To address the above problems, we introduce an end-to-end model that can leverage visual prompts to detect multi-object keypoints. For effective training, we unify the existing $13$ datasets to generalize the model across the object and their keypoints.

\subsection{Textual Prompt-based Keypoint Detection}
Benefit from vision-language pretrained model CLIP~\cite{radford2021learning},  
like open-vocabulary object detection and semantic segmentation tasks are actively explored~\cite{zang2022open,gu2021open,li2022grounded,yao2022detclip,liu2023grounding,liang2023open,zhong2022regionclip,li2023semantic,sun2023going}. In the field of keypoint detection, 
CLAMP~\cite{zhang2023clamp} is the first work to leverage CLIP with language guidance to prompt the animal keypoints containing a fixed keypoint set (e.g., 20). 
CLAMP's primary emphasis is cross-species generalization within a predefined skeleton structure. However, its limited support for keypoint descriptions and the two-stage paradigm restrict its effectiveness.
In this work, \ModelName introduces a large-scale dataset annotated with more keypoint names. Trained on such a dataset, we offer an end-to-end model that can leverage textual prompts to detect multi-object keypoints. Moreover, we are the first to explore multi-modal prompt-based keypoint detection.

%% file: Sec/Method.tex
\section{Method}
\label{sec:Overview}
\begin{figure*}[t]	
\vspace{-0.4cm}
\centering
 	{
 			\centering         
 			\includegraphics[width=1.0\linewidth]{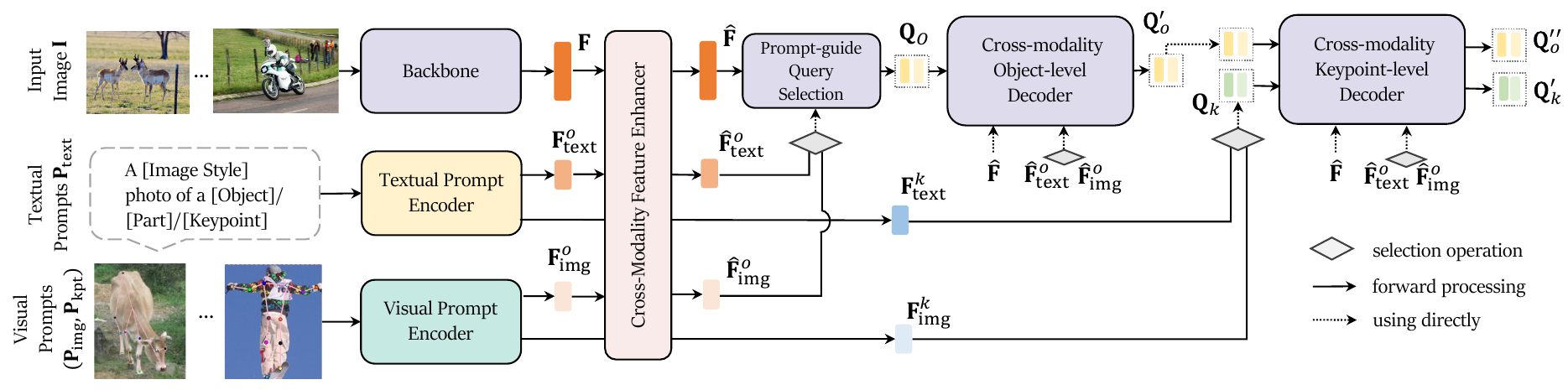}   
 	} 
\vspace{-0.3cm}
\caption{The overview architecture of \ModelName. Given an input image, \ModelName follows the coarse-to-fine strategy to detect keypoints of any object via textual or visual prompts.
}
\label{fig:framework} 
\end{figure*}

As shown in Fig.~\ref{fig:framework},
\ModelName is an end-to-end multi-modal prompt-based keypoint detection framework.
It takes an image accompanied by textual or visual prompts as input and outputs all the object bounding boxes and the corresponding keypoints. In the following subsection, we first introduce how to encode multi-modal inputs and enhance each other in Sec.~\ref{sec:encoding}. Then, we illustrate how to end-to-end decode the prompt-oriented information, including the desired object bounding boxes and keypoints in Sec.~\ref{sec:decoding}. Finally, we provide the training loss function and inference pipeline in Sec.~\ref{sec:loss}.

\begin{figure*}[t]	
\centering
 	{
 		\begin{minipage}[t]{1.0\linewidth}
 			\centering         
 			\includegraphics[width=1.0\linewidth]{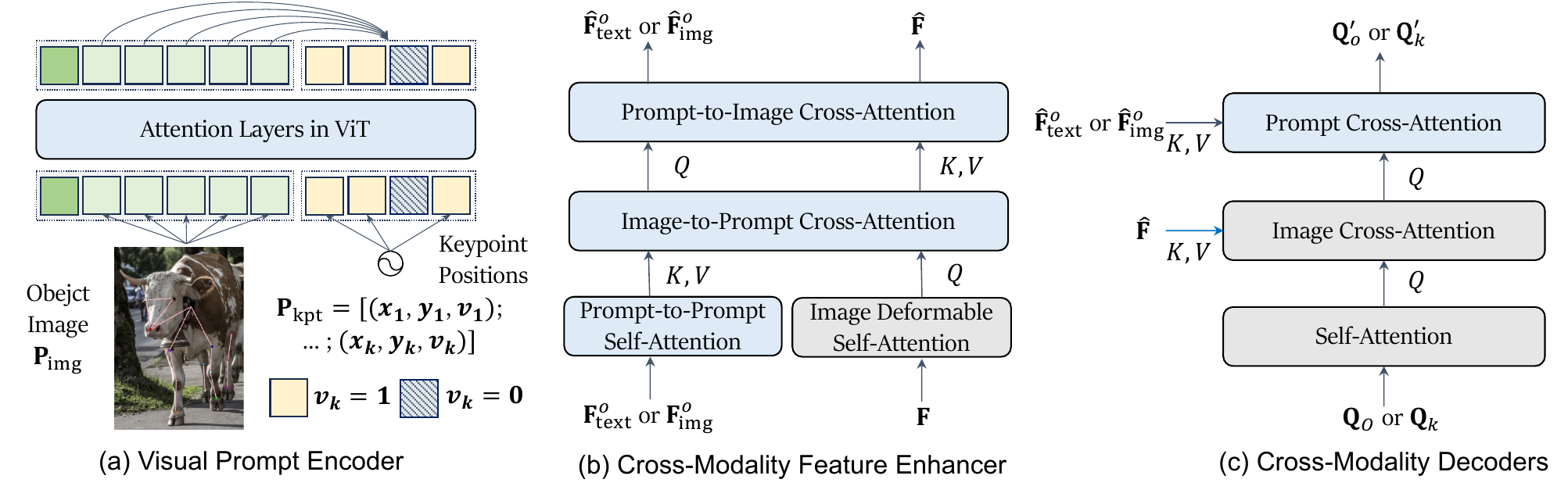}   
 		\end{minipage}
 	} 
\vspace{-0.5cm}
\caption{The detailed illustration of a) Visual Prompt Encoder, b) Cross-Modality Interactive Encoder, and c) Cross-Modality Interactive Decoder. In (b) and (c), blue modules are newly introduced to incorporate prompt interactions.
}
\label{fig:modules} 
\vspace{-0.5cm}
\end{figure*}

\subsection{Multi-Modality Inputs Encoding and Enhancing}
\label{sec:encoding}

\ModelName supports three kinds of inputs: input image $\textbf{I}$, textual prompts $\textbf{P}_{\rm{text}}$, and visual prompts $(\textbf{P}_{\rm{img}},\textbf{P}_{\rm{kpt}})$, where $\textbf{P}_{\rm{img}}$ is an image of object and $\textbf{P}_{\rm{kpt}}$ is 2D positions of defined keypoints. 
The input image is processed through a backbone network, to extract multi-scale and tokenized features $\textbf{F}$. For multi-modal prompts, we introduce two novel encoding mechanisms as follows:

\noindent \textbf{Textual Prompts} are encoded by the CLIP text encoder to produce object-level and keypoint-level textual embeddings $\textbf{F}_{\rm{text}}^{o}$ and $\textbf{F}_{\rm{text}}^{k}$, respectively. In particular, we formulate the textual prompt $\textbf{P}_{\rm{text}}$ as a hierarchical structure to describe objects, parts ({e.g.,} face and hand), and keypoints. The template can be written as: {``An {[image style]} photo of a {[object]/[part]/[keypoint]}”}.

\noindent \textbf{Visual Prompts} are processed through the CLIP image encoder to acquire visual embeddings for objects $\textbf{F}_{\rm{img}}^{o}$ and keypoints $\textbf{F}_{\rm{img}}^{k}$. As shown in Fig.~\ref{fig:modules}-(a), while the original CLIP image encoder accepts only images as input, \ModelName extends its capability to incorporate keypoints for position encodings. 
\textbf{Firstly}, given a specific keypoint $(x,y,v)$, where $(x,y)$ denotes its 2D coordinate and $v$ indicates the visibility of the keypoint, we introduce two token initialization methods for different keypoints: i) for visible keypoints ($v$=$1$), we use the Fourier embedding~\cite{mildenhall2021nerf} to map the 2D coordinate to the corresponding feature dimensions; 
ii) for invisible keypints ($v$=$0$), we employ a shared learnable mask token~\cite{he2022masked} to represent the invisible position.
\textbf{Secondly},
since the initialized keypoint tokens solely contain position information, we adopt two encoding mechanisms: i) the ``keypoint token to keypoint token” attention to capture skeletal structure relations; ii) the ``image patch token to keypoint token” attention to propagate global image feature into each keypoint token.

\noindent \textbf{Cross-Modality Enhancing.} As in Fig.~\ref{fig:modules}-(b), after obtaining the image features $\textbf{F}$ and prompt features $\textbf{F}_{\rm{text}}^o$,  $\textbf{F}_{\rm{img}}^o$, we employ the Deformable
self-attention \cite{zhu2020deformable} to enhance image features and the vanilla self-attention~\cite{vaswani2017attention} for enhancing prompt features. Moreover,
we leverage an image-to-prompt cross-attention and a prompt-to-image
cross-attention for cross-modality enhancement.

\subsection{Cross-Modality Object and Keypoint Decoding}
\label{sec:decoding}

The decoders of \ModelName are decoupled into the object-level decoder and the keypoint-level decoder. Leveraging the enhanced image features and multi-modal prompt features in Sec.~\ref{sec:encoding}, \ModelName initializes the object queries $\textbf{Q}_o$ and keypoint queries $\textbf{Q}_k$ to decode all the objects $\textbf{Q}_o^{\prime \prime}$ with their associated keypoints $\textbf{Q}_k^{\prime}$.

\noindent \textbf{Object Decoding.} {Firstly}, we utilize prompt-guided query selection~\cite{zhang2022dino,liu2023grounding} to initialize object queries $\textbf{Q}_o$ from the enhance image features $\widehat{\textbf{F}}$, which is highly associated with the enhanced object-level prompt features
$\widehat{\mathbf{F}}_{\rm{text}}^{o}$ or $\widehat{\mathbf{F}}_{\rm{img}}^{o}$. {Then}, a cross-modality object decoder is employed to update these object queries to $\textbf{Q}_o^{\prime}$. Illustrated in Fig.~\ref{fig:modules}-(c), object queries are inputted into a self-attention layer, an image cross-attention layer to integrate image features, and a prompt cross-attention layer to integrate prompt features.

\noindent \textbf{Keypoint Decoding.} The keypoint queries $\mathbf{Q}_{k}$ are directly initialized by keypoint level prompt features $\widehat{\mathbf{F}}_{\rm{text}}^k$ or $\widehat{\mathbf{F}}_{\rm{img}}^k$. {Then}, a cross-modality keypoint decoder is employed to update these keypoint queries to $\textbf{Q}_k^{\prime}$. Similar to object decoding,
keypoint queries are processed through a self-attention layer, an image cross-attention layer to combine image features, and a prompt
cross-attention layer to combine prompt features.

\subsection{Training and Inference Pipeline}
\label{sec:loss}

We adopt the same object and keypoint regression losses as previous end-to-end keypoint detectors~\cite{shi2022end,yang2022explicit,yang2023neural}: the L1 loss and the GIOU loss~\cite{rezatofighi2019generalized} for object's bounding box regression $\mathcal{L}_{reg}^{obj}$; the L1 loss and the OKS loss~\cite{shi2022end} for keypoint regression $\mathcal{L}_{reg}^{kpt}$. 
Moving forward, \ModelName introduces prompt-to-object and prompt-to-keypoint contrastive losses for alignments.

\noindent \textbf{Object-level Alignment.}
Previous keypoint detectors~\cite{xu2022vitpose,sun2019deep,yang2022explicit} mainly focus on close-set objects and typically use a simple linear layer as the object classifier. In contrast, \ModelName encodes multi-modal prompts (text or image) into the object-level prompt features 
$ \widehat{\mathbf{F}}_{\rm{text}}^{o}, \widehat{\mathbf{F}}_{\rm{img}}^{o} \in \mathbb{R}^{L\times C} $, where $L$ is the number of object classes in prompts and $C$ indicates the feature dimension. 
Following~\cite{li2022grounded,liu2023grounding}, we employ contrastive loss between predicted objects $\textbf{Q}_{o}^{\prime \prime} $ and prompt features for classification. More specifically, we compute the dot product between each object query and the prompt features to predict logits and then calculate the Focal loss of each logit $\mathcal{L}_{align}^{obj}$ for optimization.

\noindent \textbf{Keypoint-level Alignment.}
In previous keypoint detectors, the classification problem for keypoints is often overlooked, and the learning process is to establish a one-to-one mapping between predicted and labeled keypoints. In contrast, \ModelName takes the first step toward prompts-to-keypoint alignment using a unified set of keypoint definitions. Similar to object-level alignment, given keypoint prompt features $ \widehat{\mathbf{F}}_{\rm{text}}^{k}, \widehat{\mathbf{F}}_{\rm{img}}^{k} \in \mathbb{R}^{K\times C}$,
where $K$ is the number of keypoint categories in prompts. We utilize contrastive loss between predicted keypoints $\textbf{Q}_{k}^{\prime}$ and prompt features
for classification.

\noindent \textbf{The Overall Loss.} The overall training pipeline of \ModelName can be written as follows, 
\begin{equation}
   \mathcal{L} = \mathcal{L}_{reg}^{obj} + \mathcal{L}_{reg}^{kpt} + \mathcal{L}_{align}^{obj} + \mathcal{L}_{align}^{kpt} 
\end{equation}

\noindent \textbf{Training \& Inference Details}. {During training}, we employ a 50\% probability to randomly select either a visual prompt or textual prompt for each iteration. We sample two images containing the same object category from our {UniKPT} dataset and then choose one as the visual prompt. {During inference}, 
1) Textual Prompts: We can utilize pre-defined object classes with keypoints definitions as text prompts to obtain quantitative results. In practical scenarios, users can provide the text to predict the desired objects with keypoints or any keypoint.
2) Visual Prompt: We can randomly sample a set of image prompts from the training data to obtain quantitative results. In practical scenarios, users can provide a single object image (1-shot) with the corresponding keypoint definition to predict all the similar objects in the test images.

%% file: Sec/data.tex
\section{UniKPT: A Unified Keypoint Dataset}
\begin{table}[t]
    \setlength\tabcolsep{6pt}
     \caption{Statistics of {UniKPT} with 13 keypoint datasets.}
 \vspace{-1cm}
	\begin{center}
\resizebox{1\linewidth}{!}{
\begin{threeparttable}
		\begin{tabular}{c|cccc|cc}
			\hline
			Datasets & Keypoints & Class & Images & Instances & Unified Images & Unified Instances\\ \hline
            COCO~\cite{lin2014microsoft} & 17 & 1  & 58,945 & 156,165 & 58,945 & 156,165\\
            300W-Face~\cite{sagonas2016300} & 68 & 1  &  3,837 & 4,437 & 3,837 &  4,437\\
            OneHand10K~\cite{wang2018mask} & 21 & 1  & 11,703 & 11,289 & 2,000 & 2000 \\
            Human-Art~\cite{ju2023human} & 17 & 1 & 50,000 & 123,131 & 50,000 & 123,131 \\ 
            AP-10K~\cite{yu2021ap} & 17 & 54  & 10,015&  13,028  & 10,015 & 13,028 \\ 
            APT-36K~\cite{yang2022apt} & 17 & 30  & 36,000 & 53,006 & 36,000 & 53,006  \\ 
            MacaquePose~\cite{labuguen2021macaquepose} & 17 & 1 &13,083 & 16,393 & 2,000 & 2,320 \\
            Animal Kingdom~\cite{ng2022animal} & 23& 850 & 33,099  & 33,099 & 33,099 & 33,099 \\
            AnimalWeb~\cite{khan2020animalweb} & 9 & 332 & 22,451 & 21,921& 22,451 &  21,921 \\
            Vinegar Fly~\cite{pereira2019fast} & 31 & 1  &  1,500 & 1,500 & 1,500 & 1,500 \\
            Desert Locust~\cite{graving2019deepposekit} & 34 & 1  & 700 & 700 & 700 & 700  \\
            Keypoint-5~\cite{wu2016single} & 55/31\tnote{1} & 5 & 8,649 & 8,649 & 2,000 & 2,000  \\
            MP-100~\cite{xu2022pose} & 561/293\tnote{1} & 100 & 16,943 & 18,000  & 16,943 &  18,000\\  
            \hline
           UniKPT  &338 & 1237 & - & - & 226,547 & 418,487 \\ \hline
		\end{tabular}
		         \begin{tablenotes}   
        %\footnotesize  
        \tiny
        \item[1] Keypoint-5 and MP-100 have different categories with varying numbers of keypoints. 
        While the cumulative count of keypoints reaches $55$ and $561$ by aggregating across categories, we consolidate them into unified counts of $31$ and $293$ keypoints by leveraging textual descriptions.
        \item[2] MP-100 includes training subsets from two other datasets, Deepfashion2~\cite{ge2019deepfashion2} and Carfusion~\cite{reddy2018carfusion}.
      \end{tablenotes}            
    \end{threeparttable}   }
	\end{center}	
 	\label{tab:datasets}
\vspace{-0.5cm}
\end{table}

\noindent \textbf{Unifying 13 Keypoint Datasets into {UniKPT}.} As summarized in Tab.~\ref{tab:datasets}, we observe that each dataset only focuses on a single super-category (e.g., ``human only'' and ``animal only''), making it challenging to achieve keypoint generalization when using them individually. Additionally, these datasets have significant differences in quality, quantity, and appearance styles. 
Motivated by these, we propose to unify $13$ existing keypoint detection datasets into \DataName. 
which addresses several crucial aspects:
1) Balanced Diversity: We ensure a balance across the $13$ datasets by considering diverse factors such as image appearance, style, poses, viewpoints, visibilities, and scales.
2) Semantic Relationships: We reconstruct the semantic relationships between all keypoints and categories in the $13$ datasets. 
3) Standardized Definitions: We standardize the definitions of the same keypoints across different categories. For example, the left eye of various animals would be consistently defined.
4) \textbf{Enhanced Annotations}: Each keypoint within \DataName has been meticulously annotated with a unique name to enhance keypoint-level semantic understanding.

\noindent \textbf{Statistical Analysis.} In total, the unified dataset comprises $226,547$ images and $418,487$ instances, featuring $338$ keypoints and $1,237$ instance categories. In particular, for articulated objects like humans and animals, we further categorize them based on biological taxonomy, resulting in $1,216$ species, $66$ families, $23$ orders, and $7$ classes.

%% file: Sec/experiment.tex
\section{Experiment}
\subsection{Experimental Setup}
\label{Experimental_Setup}
\textbf{Dataset.} We follow the fair and standard benchmarks: 1) MP-100~\cite{xu2022pose} for visual prompt-based keypoint detection in Sec.~\ref{exp:visual}; 2) AP-10K~\cite{yu2021ap} for textual prompt-based keypoint detection in Sec.~\ref{exp:textual}; 3) \DataName for general keypoint detection in Sec.~\ref{exp:general}. Please refer to the Appendix for more details about each dataset.

\noindent \textbf{Implementation details.} 1) {{Network Details.}} We use $6$-layer cross-modality feature enhancer, $2$-layer cross-modality object decoder, and $4$-layer cross-modality keypoint decoder. We adopt the CLIP model with the ViT-Base network for visual and textual prompt encoding. The feature dimension is set to $256$. 
2) {{Training \& Inference Details.}}
We use the exact same training details as all the end-to-end models~\cite{yang2022explicit,shi2022end}. Specifically, we augment the training images through random cropping, flipping, and resizing. The shorter sides are kept within $[480,800]$, while the longer sides are less than or equal to $1333$. The size of the prompt image is set to $224$, aligning with the requirement of CLIP. We utilize the AdamW optimizer with a weight decay of 1e-4. Our models are trained on 8 Nvidia A100 GPUs with a batch size of $16$. During inference, the images are resized with shorter sides of $800$ and longer sides less than or equal to $1333$.

\subsection{Qualitative In-the-wild Test}
\begin{figure*}[!h]	
% \vspace{-0.5cm}
\centering
 	{
    \includegraphics[width=1\linewidth]{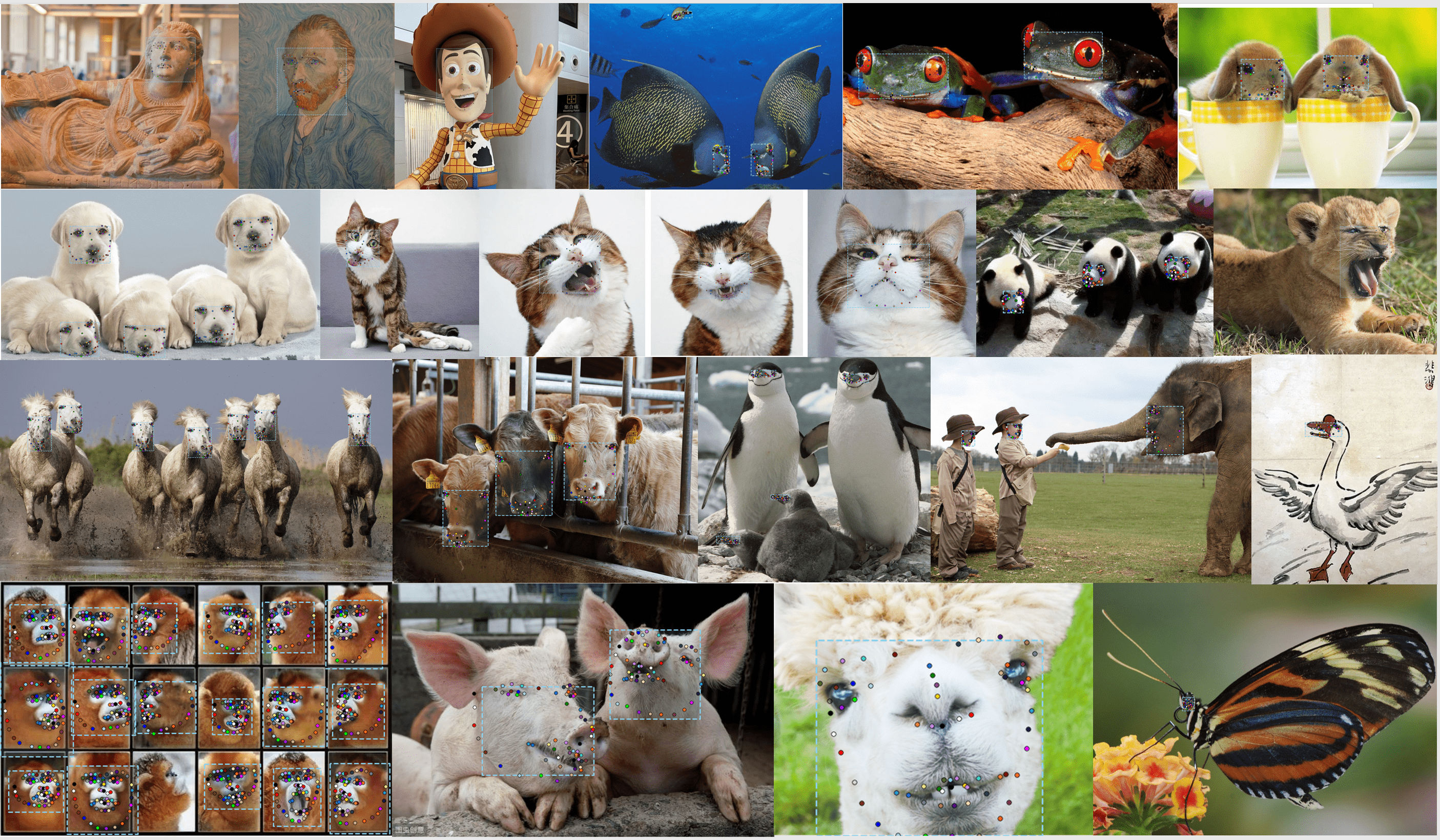}      
 	} 
\vspace{-0.5cm}
\caption{In-the-wild test of \ModelName for any face keypoint detection. We showcase the model's strong generalization to detect face keypoints of any object with 68 keypoint definitions, despite being trained only on the person's face with these definitions.}
\label{fig:any_face} 
\vspace{-0.2cm}
\end{figure*}

As shown in Fig.~\ref{fig:vis_ours}, we show the powerful detection performance of \ModelName in real-world scenarios, which could end-to-end address the challenges of cross-category (the first row), multi-object (the second row), and cross-image-style (the third row).
Furthermore, Fig.~\ref{fig:any_face} presents a surprising observation: despite being trained on a limited dataset containing human faces with the $68$ keypoints defined by~\cite{sagonas2016300}, \ModelName demonstrates remarkable cross-object capabilities when tasked with any face detection.

\subsection{Visual Prompt-based Keypoint Detection}
\label{exp:visual}
We evaluate \ModelName against the previous methods, such as ProtoNet \cite{snell2017prototypical}, MAML \cite{finn2017model}, Fine-tune~\cite{nakamura2019revisiting}, POMNet~\cite{xu2022pose}, and Capeformer~\cite{shi2023matching} on the MP-100 dataset, as shown in Tab.~\ref{tab:mp_100_main}. As an end-to-end framework, \ModelName offers efficiency by requiring only a single forward pass for scenes with multiple objects. Surprisingly, it outperforms all top-down methods, establishing a new state-of-the-art performance. 
This achievement can be attributed to \ModelName's utilization of instance-level visual prompt information, which enhances keypoint-level generalization, especially in such scenarios involving unseen objects with significant variations in appearance and different keypoint definitions.

\begin{table*}[!h]
\caption{Comparisons of visual prompt-based keypoint detection on the MP-100 benchmark. $\dag$ means the model needs keypoint identifiers during the test, however, which is not available in real-world scenarios. The inference times for all methods are tested on an A100 with a batch size of 1. Top-down methods need multiple inferences when $N$ objects are detected in an image. The compared \underline{results} are highlighted.}
\vspace{-1cm}
    \setlength\tabcolsep{6pt}
	\begin{center}
\resizebox{\linewidth}{!}{
     \begin{threeparttable}  
		\begin{tabular}{c|l|c|cccccc|c}
			\hline
			\multicolumn{2}{c|}{Method} & Backbone & Split1 & Split2 & Split3 & Split4 & Split5 & Mean (PCK) & Time [ms]\\ \hline
           \multirow{5}{*}{{\textbf{TD}}}  & ProtoNet & ResNet-50  &46.05 & 40.84 & 49.13 & 43.34 & 44.54 & 44.78 & -\\
           &  MAML & ResNet-50 & 68.14 & 54.72 & 64.19 & 63.24 & 57.20 & 61.50 & -\\
            & Fine-tune &ResNet-50& 70.60 & 57.04 & 66.06 & 65.00 & 59.20 & 63.58 & -\\
            & POMNet &ResNet-50 & 84.23 & 78.25& 78.17 & 78.68 & 79.17 & \underline{79.70} & 151$\times$ \textit{N}\\
            & CapeFormer$^\dag$ & ResNet-50  & \textbf{89.45} & 84.88 & 83.59 & 83.53 & 85.09 & 85.31 & 57$\times$ \textit{N}\\
        & CapeFormer & ResNet-50  & \underline{85.81} & - & - &  - & - &  -  & 57$\times$ \textit{N}\\
            \hline
              \multirow{1}{*}{{\textbf{E2E}}}  & \ModelName & ResNet-50  & 89.07{\color{Red}{$\uparrow_{3.26}$}} & \textbf{85.05} & \textbf{85.26} & \textbf{85.52} & \textbf{85.79} & \textbf{86.14}{\color{Red}{$\uparrow_{6.44}$}} & \textbf{59}\\
              \hline
		\end{tabular}
  \begin{tablenotes}
        \footnotesize  
\item[] Note: We train our models only on the MP-100 dataset to ensure a fair comparison.
During evaluation, all methods use the same visual prompts paired with test images.
\end{tablenotes}
  \end{threeparttable}
  }
  
       \label{tab:mp_100_main}
	\end{center}
% \vspace{-0.7cm}
\end{table*}

\subsection{Textual Prompt-based Keypoint Detection}
\label{exp:textual}
To compare with the previous textual prompt-based model-CLAMP~\cite{zhang2023clamp}, we follow its zero-shot setting to evaluate the model’s generalization ability on unseen animal species, using AP-10K. In particular, we select the Bovidae or Canidae animal orders as the training set and the unseen orders Canidae and Felidae as the testing set.
The results are shown in Tab.~\ref{tab:CLAMP}.  Compared to CLAMP, \ModelName achieves
much better performance in both settings, e.g., there is a
6.9 AP and 7.0 AP increase in these two settings, respectively. Moreover, \ModelName as an end-to-end model has greater efficiency when dealing with multi-object scenes.

\begin{table*}[!h]
    \centering
    \vspace{-0.5cm}
        \caption{Comparisons of textual prompt-based keypoint detection on AP-10K. Following CLAMP~\cite{zhang2023clamp}'s zero-shot setting, we train the model only on Bovidae or Canidae animal order and test it on unseen Canidae or Felidae animal order, respectively. }
        \vspace{-0.2cm}
    \resizebox{0.8\linewidth}{!}{
        \begin{threeparttable}
            \begin{tabular}{l|c|cc|ccc|c}
                \hline
                Methods & Backbone & Train & Test & ${\rm AP}$ &  ${\rm AP}_{M}$ & ${\rm AP}_{L} $ & Time [ms] \\ \hline
                CLAMP (TD) & ResNet50 & Bovidae & Canidae & 46.9 & 30.3 & 46.9  & 63 $\times$ \textit{N} \\
                \ModelName (E2E) & ResNet50 & Bovidae & Canidae & \textbf{53.8}{\color{Red}{$\uparrow_{6.9}$}} & \textbf{34.7} & \textbf{54.1} & \textbf{59} \\ \hline
                CLAMP (TD) & ResNet50 & Canidae & Felidae & 48.4 & 13.6  & 48.9 & 63 $\times$ \textit{N} \\
                \ModelName (E2E) & ResNet50 & Canidae & Felidae & \textbf{55.4}{\color{Red}{$\uparrow_{7.0}$}} & \textbf{20.8} & \textbf{54.2} & \textbf{59}\\ \hline
            \end{tabular}
        \end{threeparttable}
    }
    \label{tab:CLAMP}
    \vspace{-0.2cm}
\end{table*}

\subsection{General Keypoint Detection}
\label{exp:general}

\noindent \textbf{Comparison with the SOTA on Various Keypoint Datasets.}
Existing state-of-the-art results are achieved by non-promptable models with two-stage schemes. Compared to them, \ModelName possesses a unique advantage in superior generalization to handle unseen objects with different keypoint definitions. Furthermore, as shown in Tab.~\ref{tab:sota_comparison}, \ModelName surprisingly achieves new state-of-the-art performance on most datasets. Despite slightly lower performance on some datasets like COCO, \ModelName demonstrates for the first time the performance upper ceiling of an end-to-end model.

\noindent \textbf{Comparison with the SOTA End-to-End Model.}
For a fair comparison, we train both our \ModelName and ED-Pose using the same datasets, \textit{i.e.}, COCO, Human-Art, AP-10K, and APT-36K. The complete results are shown in the Appendix. Tab.~\ref{tab:comp_edpose} highlights the performance comparison on AP-10K, which involves the classification of 54 different species. \ModelName surpasses ED-Pose with a $27.7$ AP improvement, thanks to instance-level and keypoint-level alignments.

\begin{figure}[h]	
\vspace{-0.2cm}
\centering
 	{
 			\centering         
 			\includegraphics[width=1.0\linewidth]{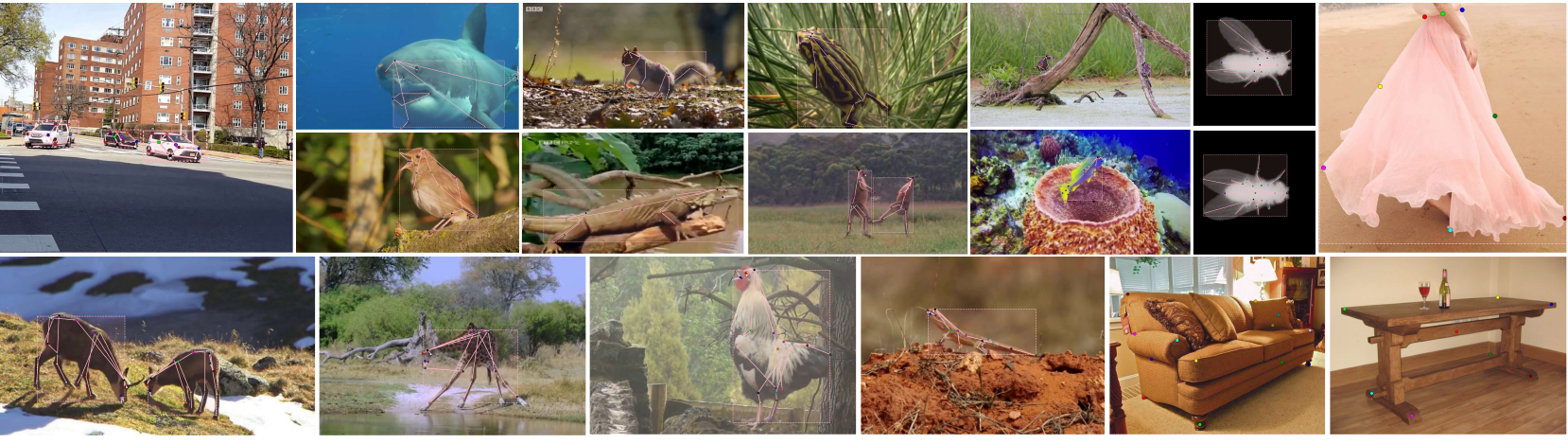}   
 	} 
\vspace{-0.5cm}
\caption{Visualization of the detected keypoints via \ModelName on {UniKPT}.
}
\label{fig:vis_dataset} 
\vspace{-0.3cm}
\end{figure}

\noindent \textbf{Qualitative Results on Existing Datasets.} Given an input image and textual prompts, \ModelName shows powerful qualitative results across existing closed datasets, encompassing articulated, rigid, and soft objects, as shown in Fig.~\ref{fig:vis_dataset}.

\begin{table*}[t]
    \setlength\tabcolsep{6pt}
    	\caption{Comparison with absolute SOTA results on all existing keypoint datasets. $\dag$ indicates results using the flipping test. Results marked with * rely on ground-truth bounding boxes for top-down methods. The \textbf{best} results are highlighted in \textbf{bold}, and the \underline{second best} results are highlighted with a \underline{underline}. \textit{T} and \textit{V} denote textual and visual prompts used.}
     \vspace{-0.8cm}
	\begin{center}
\resizebox{\linewidth}{!}{
\begin{threeparttable}
		\begin{tabular}{l|c|cccccccccccc}
			\hline
			\multirow{2}{*}{Methods} & \multirow{2}{*}{Backbone} & COCO & AP-10K & Human-Art & \multicolumn{1}{c|}{Macaque}  & 300W & Hand & AK & Fly & Locust & KPT-5 & DF2 & Carfusion \\ 
    & & \multicolumn{4}{c|}{AP$\uparrow$}   &\multicolumn{8}{c}{PCK$\uparrow$} \\ \hline
                        SOTA (TD)  & -  & \cellcolor{blue!10} \textbf{78.6}$^\dag$  & \cellcolor{blue!10} \textbf{80.4*}$^\dag$ & \cellcolor{blue!5} 35.6 & \cellcolor{blue!5} 51.9*  & \cellcolor{blue!5} \textbf{99.8*} & \cellcolor{blue!5} 99.5* & - & - & - & -& - & -  \\ \hline
                        \ModelName-\textit{T} (E2E)& Swin-T & 74.4  & 74.0  & 72.5 & 78.0 & 98.1  & 95.7 & 95.3 & 99.6 & 99.7  & 94.3 & 95.7 & 78.1\\
                        \ModelName-\textit{V} (E2E)& Swin-T& 74.3 & 73.6 & 72.1  & 77.3 &  \underline{99.4} & 95.9 &  94.3 & 99.8  & 99.6 & 87.4 & 91.0 & 72.1 \\
                        \ModelName-\textit{T} (E2E)& Swin-L &  \underline{76.8} & \underline{79.2}  & \textbf{75.9}  & \textbf{79.4}  & 98.5 & \textbf{99.8} & \textbf{96.1} & \textbf{99.9} & \textbf{99.8} & \textbf{95.5} & \textbf{97.5} & \textbf{88.7} \\
                        \ModelName-\textit{V} (E2E)& Swin-L &  76.6  & 79.0 & 75.5  & 77.8  &99.3 & 99.5& 95.5 & \textbf{99.9} & \textbf{99.9} & 91.6 & 95.5 & 85.0 \\ \hline
                        
		\end{tabular}     
    \end{threeparttable}   
    }
 \label{tab:sota_comparison}
	\end{center}
\vspace{-0.3cm}
\end{table*}

\begin{table*}[t]
    \begin{center}
                            \begin{minipage}{0.45\linewidth}
                            	\caption{Comparisons with the end-to-end non-prompted model on AP-10K. We train ED-Pose with the same dataset as \ModelName.}
                             \vspace{-0.3cm}
                            \centering
        \resizebox{\linewidth}{!}{
            \makeatletter\def\@captype{table}\makeatother
		\begin{tabular}{l|c|ccc}
    \hline
Methods  & Backbone    & ${\rm AP}$ &  ${\rm AP}_{M}$ & ${\rm AP}_{L}$ \\ \hline
               ED-Pose & Swin-T& \underline{45.5}  & 31.0 & 46.5  \\
              \ModelName-\textit{V} & Swin-T   & 72.8  & \textbf{47.2} & 74.0   \\
                \ModelName-\textit{T}  & Swin-T & \textbf{73.2}{\color{Red}{$\uparrow_{27.7}$}} & 45.6 & \textbf{74.3} \\
             \hline
		\end{tabular}
            }
 % \vspace{-0.3cm}
\label{tab:comp_edpose}
        \end{minipage}
            \hfill
                \begin{minipage}{0.53\linewidth}
                \caption{Comparisons of CLIP score.}
                \vspace{-0.3cm}
        \resizebox{\linewidth}{!}{
            \makeatletter\def\@captype{table}\makeatother
		\begin{tabular}{l|cc|cc}
    \hline
\multirow{2}{*}{Methods}& \multicolumn{2}{c|}{AP-10K \texttt{val}}    & \multicolumn{2}{c}{Human-Art \texttt{val}} \\ 
& Instance & Keypoint & Instance & Keypoint \\\hline
CLIP & 28.36 & 21.75 & 23.60 & 23.81\\
\ModelName & \textbf{58.59}{\color{Red}{$\uparrow_{106\%}$}}  & \textbf{66.01}{\color{Red}{$\uparrow_{204\%}$}} & \textbf{68.41}{\color{Red}{$\uparrow_{190\%}$}} & \textbf{63.46}{\color{Red}{$\uparrow_{166\%}$}}\\ 

             \hline
		\end{tabular}
            }
    \label{clip_score}
        \end{minipage}

 \vspace{-0.5cm}
    \end{center}
\end{table*}

\begin{table*}[t]
    \setlength\tabcolsep{6pt}
	\caption{Comparisons with the state-of-the-art open-vocabulary object detector, focusing on instance-level and keypoint-level detection. $\ddag$ denotes the fine-tuning of GroundingDINO using the keypoint detection datasets. Note that we limit the instance-level comparison to $AP_{M}$ (medium objects) and $AP_{L}$ (large objects), as small objects do not have keypoints annotated.}
  \vspace{-1cm}
	\label{tab:comparison_gd}
	\begin{center}
\resizebox{\linewidth}{!}{
     % \begin{threeparttable}   
		\begin{tabular}{l|c|cc|ccc|c|c}
			\hline
                        \multirow{2}{*}{Methods} &  \multirow{2}{*}{Backbone}    &   \multicolumn{2}{c|}{Instance-level} & \multicolumn{3}{c|}{Keypoint-level} & \multirow{2}{*}{Training Datasets} & \multirow{2}{*}{Dataset Volume} \\
			  &  & ${\rm AP}_{M}$ & ${\rm AP}_{L}$ &${\rm AP}$ &  ${\rm AP}_{M}$ & ${\rm AP}_{L}$  &  & \\ \hline
                \multicolumn{9}{l}{\cellcolor{Gray!20} \textit{ COCO \texttt{val} set}} \\
               GroundingDINO-\textit{T} & Swin-T & 70.8 & 82.0 &  3.1  & 2.8 & 3.2 & O365,GoldG,Cap4M & 1858K \\
                GroundingDINO-\textit{T} & Swin-B & 69.7 & 79.5 & 6.8 &  6.6 & 7.2  & COCO,O365,GoldG,Cap4M,OpenImage,ODinW-35,RefCOCO &  -\\ 
                GroundingDINO$\ddag$-\textit{T} & Swin-T & \textbf{71.2} & \textbf{83.4} & 1.8  & 1.7 & 1.9 & COCO,Human-Art,AP-10K,APT-36K & 1858K  +  155K\\
                              \ModelName-\textit{T} & Swin-T & 71.1 & 80.2 & \textbf{74.2} & \textbf{68.8}  & \textbf{82.1} & COCO,Human-Art,AP-10K,APT-36K & 155K  \\
                \ModelName-\textit{V} & Swin-T & 71.1 & 80.3 & 74.1 & \textbf{68.8} & 81.8  & COCO,Human-Art,AP-10K,APT-36K & 155K  \\
             \hline
		\end{tabular}}
		 \vspace{-0.5cm}
	\end{center}
\end{table*}

\subsection{Compared with Open-Vocabulary Models}
\textbf{Comparison with the Vision-Language Model.} 
We assess \ModelName's text-to-image alignment capabilities at different granularities, \text{i.e.}, object and keypoint. In Tab.~\ref{clip_score}, we report the CLIP score of \ModelName and CLIP~\cite{clip} on AP-10K, which involves $54$ animal categories, and Human-Art, which features $15$ image styles. Results show that \ModelName consistently provides higher-quality text-to-image similarity scores at the object and keypoint levels.

\noindent \textbf{Comparison with Open-Vocabulary Detection Model.} We compare \ModelName with the state-of-the-art open-vocabulary object detector, GroundingDINO 
\cite{liu2023grounding}, in terms of instance-level and keypoint-level detection. We present the COCO results in Tab.~\ref{tab:comparison_gd}, while results for other datasets are provided in the Appendix. Grounding-DINO fails to localize fine-grained keypoints; however, \ModelName successfully addresses these challenges, achieving significant improvements across all datasets. \ModelName maintains comparable performance with GroundingDINO. Additionally, we find that although fine-tuning GroundingDINO for instance detection can be beneficial, it negatively impacts keypoint detection.

\begin{table*}[t]
    \begin{center}
                            \begin{minipage}{0.45\linewidth}
                            \centering
                                        \caption{Impact of constrastive loss on AP-10K.}
                                        \vspace{-0.3cm}
        \resizebox{\linewidth}{!}{
            \makeatletter\def\@captype{table}\makeatother
\begin{tabular}{cc|cc|ccc}
\hline
\multirow{2}{*}{$\mathcal{L}_{Align}^{obj}$} &  \multirow{2}{*}{$\mathcal{L}_{Align}^{kpt}$}    &   \multicolumn{2}{c|}{Object-level} & \multicolumn{3}{c}{Keypoint-level}     \\  &  & ${\rm AP}_{M}$ & ${\rm AP}_{L}$ &${\rm AP}$ &  ${\rm AP}_{M}$ & ${\rm AP}_{L}$  \\ \hline
&  &  53.7& 62.5 & 45.5  & 31.0 & 46.5   \\
$\checkmark$ &  & 53.8 & 78.5 &72.6 &  43.6 & 73.4\\
$\checkmark$ & $\checkmark$ & \textbf{54.5} & \textbf{78.8} & \textbf{73.2} & \textbf{45.6} & \textbf{74.3} \\
\hline
\end{tabular}
            }
 % \vspace{-0.3cm}
\label{tab:ablation_alignment}
        \end{minipage}
            \hfill
                \begin{minipage}{0.5\linewidth}
                    \caption{Impact of multi-modal prompts on AP-10K. The prompt used in the test is highlighted in gray.}
                    \vspace{-0.3cm}
        \resizebox{\linewidth}{!}{
            \makeatletter\def\@captype{table}\makeatother
\begin{tabular}{cc|cc|ccc}
\hline
\multirow{2}{*}{Visual Prompt} &  \multirow{2}{*}{Textual Prompt}    &   \multicolumn{2}{c|}{Object-level} & \multicolumn{3}{c}{Keypoint-level}    \\  &  & ${\rm AP}_{M}$ & ${\rm AP}_{L}$ &${\rm AP}$ &  ${\rm AP}_{M}$ & ${\rm AP}_{L}$  \\ \hline
\cellcolor{gray!10} $\checkmark$ & & 53.3 & 78.1 &71.5 &  43.4 & 72.4 \\
\cellcolor{gray!10}$\checkmark$ & $\checkmark$ & \textbf{55.8} & \textbf{79.0} & 72.8 & \textbf{47.2} &74.0  \\ \hline
& \cellcolor{gray!10}$\checkmark$ & 53.8 & 78.5 & 72.9 & 45.1 & 74.2\\
$\checkmark$ & \cellcolor{gray!10}$\checkmark$ &  54.5& 78.8 & \textbf{73.2} & 45.6 & \textbf{74.3} \\
\hline
\end{tabular}
            }
    \label{ablation_prompt}
        \end{minipage}

 \vspace{-0.6cm}
    \end{center}
\end{table*}

\begin{table*}[!t]
\centering
\caption{Impact of dataset quantity on AP-10K and AnimalPose.}
\vspace{-0.3cm}
\resizebox{1\linewidth}{!}{
\begin{tabular}{c|cc|ccc|c}
\hline  \multirow{2}{*}{Training Data}    &   \multicolumn{2}{c|}{ AP-10K's Object} & \multicolumn{3}{c|}{AP-10K's Keypoint}   & AnimalPose  \\  & ${\rm AP}_{M}$ & ${\rm AP}_{L}$ &${\rm AP}$ &  ${\rm AP}_{M}$ & ${\rm AP}_{L}$  & PCK\\ \hline
COCO,Human-Art,AP-10K,APT-36K & 54.5& 78.8 & 73.2 & 45.6 & 74.3 & 52.7\\
+MacquePose,AnimalKingdom,AnimalWeb,Vinegar Fly,Desert Locust& \textbf{55.6} & \textbf{80.2} & \textbf{74.2} & \textbf{48.3} & \textbf{75.0} & 70.1\\
+300w-Face,OneHand10K,Keypoint-5,MP-100 &  55.3 & 78.8  & 74.0 & 47.8 & 74.7 & \textbf{73.4} \\
\hline
\end{tabular}}
 \vspace{-0.3cm}
\label{ablation_data}
\end{table*}

\subsection{Ablation Study}

In the first two ablation studies, we train \ModelName with the Swin-T backbone on four datasets: COCO, Human-Art, AP-10K, and APT36K. For fair comparisons, we report the results on AP-10K, which enables comprehensive evaluation in classification and localization. In the third ablation study, we present the results on both the seen dataset AP-10K in {UniKPT} and the unseen dataset AnimalPose~\cite{cao2019cross} to demonstrate generalization ability.

\myPara{Contrastive Loss.} We introduce $\mathcal{L}_{Align}^{obj}$ and $\mathcal{L}_{Align}^{kpt}$ to facilitate prompt-to-object and prompt-to-keypoint alignment, respectively, as in Sec.~\ref{sec:loss}. We present the results using textual prompts in Tab.~\ref{tab:ablation_alignment}, highlighting the significant improvement in detection performance, particularly in $AP_L$, due to $\mathcal{L}_{Align}^{obj}$. This underscores its importance to benefit the model to distinguish between categories and enhance classification performance.
The improved detection performance positively affects keypoint performance. Moreover, the inclusion of $\mathcal{L}_{Align}^{kpt}$ further helps the network learn keypoint distinctions, resulting in enhanced keypoint detection performance.

\myPara{Multi-Modality Prompts.} 
We explore whether the two modalities can benefit each other in Tab.~\ref{ablation_prompt}. Single-modality training settings always perform worse than multi-modality settings, highlighting the mutual advantages of both textual and visual prompts.

\myPara{Dataset Quantity in {UniKPT}.} We first train our \ModelName using $4$ datasets covering humans and $60$ different animals. Then, we add extra $5$ animal datasets to train \ModelName, as shown in Tab.~\ref{ablation_data}. This results in significant improvements in both instance and keypoint detection on seen AP-10K datasets (using textual prompts). Moreover, we achieve a significant improvement on the unseen AnimalPose dataset (using visual prompts), thanks to more categories and the increased data size, making the model more generalizable. Furthermore, we incorporate additional part-level datasets (Face and Hand) as well as rigid and soft object datasets for training. Although these diverse datasets lead to a slight decrease in AP-10K performance, it further boosts the model's performance on unseen datasets.

\subsection{The Analysis of Multi-modal Prompts}
% \noindent \textbf{Flexibility of Multi-modal Prompts.}
% In general, textual prompts are efficient for describing most object and keypoint categories, offering a more effective user interaction solution. However, when objects and keypoints are challenging to describe textually, visual prompts can enhance effectiveness.
% As in Fig.~\ref{fig:fig2}-(d), we show the flexibility of our model to support visual prompts only, textual prompts only, a combination of visual and textual prompts, and scenarios where no prompt input is required (utilizing predefined categories in UniKPT for inference). The results underscore the powerful detection performance and flexibility of \ModelName in real-world applications.

\begin{figure*}[h]	
\vspace{-0.3cm}
\centering
 	{
 			\centering         
 			\includegraphics[width=1\linewidth]{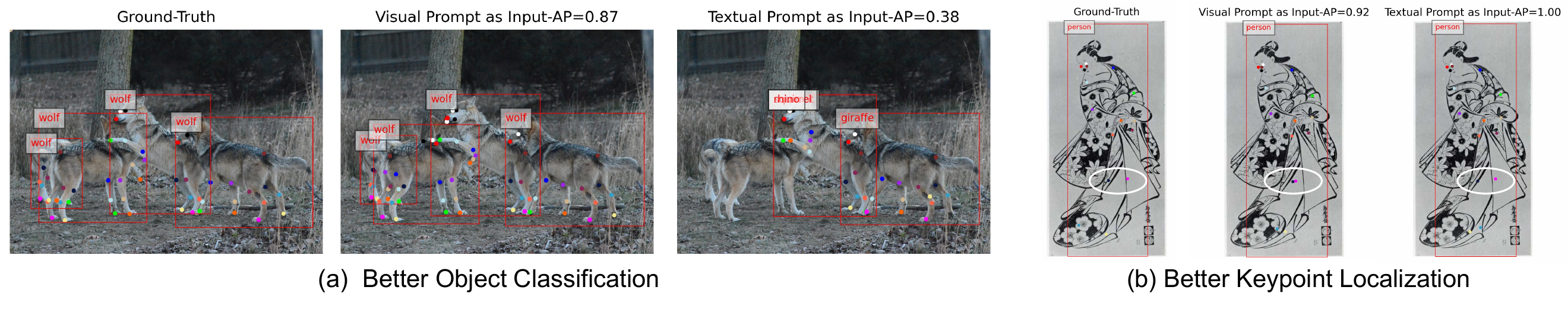}   
 	} 
\vspace{-0.6cm}
\caption{Prompts Analysis in \ModelName on (a) AP-10K~\cite{yu2021ap} and (b) Human-Art~\cite{ju2023human}. 
}
\label{fig:prompt_analysis} 
\vspace{-0.6cm}
\end{figure*}

% \noindent \textbf{Advantages of Different Prompts.}
As in Fig.~\ref{fig:prompt_analysis}, we present a visualization comparison of using different prompts across AP-10K and Human-Art datasets. We make two key observations: i) Since AP-10K requires accurate animal category classification (e.g., from 54 categories), we find that the classification accuracy achieved through visual prompts surpasses that of textual prompts (see Fig.~\ref{fig:prompt_analysis}-(a)). This is attributed to the fact that visual prompts can provide a greater volume of similar instance features, enhancing the accuracy of classifications.
ii) We notice that text prompts offer a slightly better keypoint localization accuracy compared to visual prompts (see Fig.~\ref{fig:prompt_analysis}-(b)). However, this difference is exceedingly small, primarily due to the effective cross-modality contrastive learning strategies, which significantly enhance keypoint localization accuracy.

%% file: Sec/conclusion.tex
\section{Conclusion}
This work studies the problem of detecting any keypoints in real-world scenes. To solve this problem, we proposed an end-to-end multi-modal prompt-based framework trained on a unified keypoint dataset to learn general semantic fine-grained keypoint concepts and global-to-local keypoint structure. The extensive experiments and in-the-wild tests demonstrate that \ModelName achieves high keypoint detection performance and generalizability in real-world scenes. 

\noindent  \textbf{Broader Impact:}
Based on the proposed \ModelName, we can provide 1) an end-to-end keypoint detector for any keypoints to benefit various downstream areas~\cite{yang2023semantic,yang2024open,yang2023boosting,yang2022toward,chen2024motionllm};
2) a user-friendly connector with either textual prompts or visual prompts to first detect keypoints and then take them as user clicks for fine-grained detection, segmentation, and tracking~\cite{zou2024segment,ren2024grounded,li2024taptr,kirillov2023segment}; 3) the proposed \DataName dataset could benefit the training of large vision models and its keypoint-level semantic annotations could promote better fine-grained vision-language understanding~\cite{jiang2023t,jiang2024t,wang2024visionllm,chen2024internvl}.

\section*{Acknowledgement}
The work is partially supported by the Young Scientists Fund of the National Natural Science Foundation of China under grant No.62106154, by the Natural Science Foundation of Guangdong Province, China (General Program) under grant No.2022A1515011524, and by Shenzhen Science and Technology Program JCYJ20220818103001002, and by the Guangdong Provincial Key Laboratory of Big Data Computing, The Chinese University of Hong Kong (Shenzhen).